# Teraflop-scale Incremental Machine Learning

Eray Özkural

Bilkent University Computer Engineering Department
Ankara, Turkey

**Abstract.** We propose a long-term memory design for artificial general intelligence based on Solomonoff's incremental machine learning methods. We use R5RS Scheme and its standard library with a few omissions as the reference machine. We introduce a Levin Search variant based on stochastic Context Free Grammar together with four synergistic update algorithms that use the same grammar as a guiding probability distribution of programs. The update algorithms include adjusting production probabilities, re-using previous solutions, learning programming idioms and discovery of frequent subprograms. Experiments with two training sequences demonstrate that our approach to incremental learning is effective.

## 1 Introduction

Artificial General Intelligence (AGI) field has received considerable attention from researchers in the last decade, as the computing capacity marches towards human-scale. Many promising theoretical proposals have been put forward [1,2,3] and practical general-purpose programs have been demonstrated (for instance [4,5]). We currently understand the requirements of an AGI system much better than we used to, therefore we believe that it is now time to start constructing a complete AGI system. Thus, we set out to expose and meet the experimental challenges early on.

Teramachine is our prototype implementation of an AGI system in the O'Caml language, as a candidate for Solomonoff's "Phase 1 machine" that he proposed to use as the basis of a powerful AGI system [6]. We report our ongoing research to share our experience in designing teramachine. Due to space restrictions we cannot give much background, and we proceed directly to our contributions. The reader is referred to [4,6,7] for a background on general-purpose incremental machine learning.

## 2 Incremental learning: Heuristic Algorithmic Memory

Solomonoff has pointed out three open problems: i) the determination of a reference machine, ii) update algorithms, iii) designing appropriate training sequences. Our approach is a (partial) solution to the first two problems.

Teramachine is a universal induction system that features integrated memory. The memory is automatic; it is recalled appropriately during induction, and after each induction problem, the solution is stored in the memory. The memory is designed around Solomonoff's idea of maintaining a "guiding pdf" of programs. The present system may also be viewed as an extension of OOPS [4], in similar vein to Adaptive Levin Search [8]. We update the guiding pdf after each induction problem so that the *heuristic* solutions that we invent are stored as *algorithmic* information in our *memory* system.

The typical use of such a transfer learning machine is to let it run in a domain, and acquire enough expertise to solve the later problems intelligently. Note that without transfer learning, the intelligence of our machine is severely handicapped. The Conceptual Jump Size (CJS) $= ti/pi$ of a solution $s_i$ is roughly the time required to find a particular solution $s_i$ that runs in $t_i$ time with a priori probability $p_i$ [6]. If the system's probability distribution of programs is fixed (i.e., no long-term memory), then the system can only find solutions with very small complexity, since there is a practical upper bound to CJS (e.g., a year). Transfer learning solves this problem by changing the probability distribution so that the already invented solutions are more likely. Our system implicitly achieves *algorithmic compression* of the solution corpus, to avoid redundant searches for algorithmic information contained in previous solutions. Therefore, the whole process can be seen as the (incremental) algorithmic compression of the the solution corpus. With that in mind, Heuristic Algorithmic Memory (HAM) has been designed to cope with very long training sequences, and extract a good deal of algorithmic information from the solution corpus. The more regularity we find in the corpus, the better chance we have to accelerate future solutions.

It may be thought that updating the probability distribution this way may hinder the appealing properties of Solomonoff induction and Levin Search. However, this is not the case. Modifying the probability distribution essentially defines an *implicit program code*. Thus, after every solution we are implicitly modifying the reference machine. As long as the reference machine maintains its universality, its desirable theoretical properties will persist. Relative to the implicit universal code, Levin search still has an optimal order of complexity, and it is an effective method to approximate Solomonoff induction. After solving each problem, we will have implicitly arranged a new reference machine that knows more about the problem domain. This corresponds to the fact that in the updated implicit program code, previously complex solutions will attain higher probability (or conversely shorter codes) and more complex potential solutions (with respect to the initial distribution) will fit into the same search time.

Therefore, the extraction of algorithmic information affords an effective kind of time-space tradeoff, which works extremely favorably in terms of additional space requirement after each update, as the successful extraction of each bit of mutual algorithmic information among two problems may potentially result in a speed-up of two for the latter problem. Bear in mind however that re-using algorithmic information from previous solutions entails a coding cost which manifests itself as a time penalty during program search.

The more intelligent and complete kinds of information extraction naturally use more time and long-term space, but eventually save more time and enable the system to cope with problems that are otherwise infeasible. In our present system, we have four fast update algorithms that modify a Stochastic Context Free Grammar of programs, which is the explicit long-term memory representation of our HAM design.

## 3 Scheme as the reference machine

For a general purpose machine learning system, we need a general purpose programming system that can deal with a large variety of data structures and makes it possible to write sophisticated programs of any kind. While FORTH has yielded rather impressive results [4], we have chosen R5RS Scheme on the grounds that it is a simple yet general purpose high-level programming language. Certain other features of it also make it desirable. It is an improvement over LISP in that it is statically scoped and its implementations are required to have proper tail recursion; it is defined precisely in a standards document [9]. R5RS contains a reasonably sized standard library. We do not think that Scheme has any major handicaps compared to Solomonoff's AZ system [7]. The small syntactic differences are not important, but language features are. Scheme does include a functional language, in addition to imperative features. It is highly orthogonal as it is built around symbolic expressions. The syntax-semantics mapping is quite regular, hence detecting patterns in syntax helps detecting patterns in semantics. There are a lot of efficient interpreters for Scheme, which may be modified easily for our uses.

We have implemented most of the R5RS syntax, with a few omissions. We have elected to exclude the syntax for quasi-quotations and syntax transformation syntax, as the advanced macro syntax would complicate our grammar based guidance logic, and as it is an advanced feature that is used only in more complex programs. Macro definition generation has been disabled altogether, as well. All of the R5RS standard library has been implemented except for input/output (6.6) and system interface (6.6.4) forming an adequate basis for generating simple programs. A special non-terminal called standard-procedure was added to the grammar which produces standard library procedure calls, with the correct number of arguments. The standard-procedure is added as an alternative production of the procedure-call head in the Scheme standard grammar. The addition of a standard library must not confuse the reader. We intentionally make the search space large to show what issues arise in a realistic system. It is completely unnecessary for the system to discover such primitive functions on its own. However, we must be able to present to it every function it needs. This is in contrast to systems that use a small number of primitives.

For generating integer literals, we employ the Zeta distribution with the pmf given by

$$P_s(k) = k^{-s}/\zeta(s)$$

where $\zeta(s)$ is the Riemann zeta function. We have used the Zeta distribution with $s = 2$ and used a pre-computed table to generate up to a fixed integer (256 in our current implementation). For variables, a robotic variable name is generated in the form of varinteger where the non-terminal integer is sampled from the first 7 values. Unbound references are avoided during generation.

## 4 Using a stochastic CFG in Levin Search

In many AGI systems, a variant or extension of Levin Search [10] is used for finding solutions. Solomonoff's incremental machine learning also uses Levin Search as the basic search algorithm to find solutions [1]. In our system, we take advantage of the stochastic grammar based guiding probability mass function (pmf) for the search procedure as well. A stochastic CFG is a CFG augmented by a probability value on each production. For each head non-terminal, the probabilities of productions of that head must sum to one, obviously.

We can extend our Levin Search procedure to work with a stochastic CFG that assigns probabilities to each sentence in the language. For this, we need two things, first a generation logic for individual sentences, and second a search strategy to enumerate the sentences that meet the termination condition of LSearch [4]. In the present system, we use leftmost derivation to generate a sentence, intermediate steps are thus left-sentential forms [11, Chapter 5]. The calculation of the a priori probability of a sentence depends on the obvious fact that in a derivation $S \Rightarrow \alpha_1 \Rightarrow \alpha_2 \Rightarrow ... \Rightarrow \alpha_n$ where productions $p_1, p_2, ..., p_n$ have been applied in order to start symbol $S$, the probability of the sentence $\alpha_n$ is naturally $P(\alpha_n) = \prod_{1 \leq i \leq n} p_i$. Note that the productions in a derivation are conditionally independent. While this makes it much easier for us to calculate probabilities of sentential forms, it limits the expressive power of the probability distribution.

A relevant optimization here is starting not from the absolute start symbol (in the case of R5RS Scheme program) but from any arbitrary sentential form. This helps fixing known parts of the program that is searched, and we have done so in the implementation.

The search strategy is important for efficient and correct implementation of the program generate-and-test step of LSearch. In our implementation, we first generate top-level sentential forms with high-probability, statically distribute them to processors and then run a probability-limited depth-first search starting from each top-level sentential form on each processor. We make use of a dynamic depth-limit in the form of a "probability horizon" which is a threshold we impose corresponding to the smallest probability sentence that we are willing to generate. The probability horizon can be calculated from the current time limit $t$ and the time quantum $t_q$ as $p_h = t_q/t$, which ensures that we will not waste time generating any programs that we will not run. Note that parallel search algorithm details and variants are beyond the scope of this paper, but we intend to address them in future publications.

## 5 Stochastic CFG updates

The most critical part of our design is updating the stochastic CFG so that the discovered solutions in a training sequence will be more probable when searching for subsequent problems. We propose four synergistic update algorithms for HAM. Our SCFG structure extends the usual productions with production procedures, which dynamically generate productions.

### 5.1 Modifying production probabilities

The simplest kind of update is modifying the probabilities as new solutions are added to the solution corpus. For this, however, the search algorithm must supply the derivation that led to the solution (which we do), or the solution must be parsed using the same grammar. Then, the probability for each production $A \to \beta$ in the solution corpus can be easily calculated by the ratio of frequency of productions $A \to \beta$ in the solution corpus to the frequency of productions in the corpus with a head of $A$. The production procedures are naturally excluded from the update as they can be variant. However, we cannot simply write the probabilities calculated this way over the initial probabilities, as initially there will be few solutions, and most probabilities will be zero. We use exponential smoothing to solve this problem.

$$s_0 = p_0$$
$$s_t = \alpha p_t + (1 - \alpha) s_{t-1}$$

where $p_0$ is the initial probability, $p_t$ is the probability in the corpus and $\alpha$ is the smoothing factor. We used a smoothing factor of 0.125. See [12] for the application of smoothing in a similar problem. Other methods like Laplace's rule may be used to avoid zero probabilities [6].

While modifying production probabilities is a useful idea, it cannot add much information to the guiding pmf as the total amount of information is limited by the number of bits per probability multiplied by the number of probabilities. While we do use arbitrary precision floating point numbers, it does not seem likely that distinguishing more finely among a few number of alternative productions for a non-terminal will result in great improvements. Then, it seems that we need to augment the grammar with new productions.

Note that the first update algorithm may be considered as a generalization of the instruction probability bumping of OOPS, since the SCFG is a generalization of the simpler program probability model used in OOPS (i.e., a program's probability is the product of the probabilities of the instructions it contains[4]), and the probability update in OOPS is dynamically caused by the bump instruction, the probability distribution is not stored long-term as in teramachine.

### 5.2 Re-using previous solutions

In the course of a training sequence, the solutions can be incorporated in full by adding the solutions to the grammar. In the case of Scheme, there could be many possible implementations. The simplest design is to add all the solutions to the library of the Scheme interpreter, add a hook non-terminal `previous-solution` to the grammar, and then extend the `previous-solution` with the syntax to call the new solution. We assume that this syntax is provided in the problem definition. We add new solutions as follows, the new solution among other previous solutions is given a probability of $\gamma$ in the hope that this solution will be re-used soon, and then the probabilities of the old productions of previous-solution are normalized so that they sum to $1 - \gamma$. We currently use a $\gamma$ of 0.5.

If it is impossible or difficult to add the solutions to the Scheme interpreter as in our case, then all the solutions can be added as `define` blocks in the beginning of the program produced. The R5RS Scheme, being an orthogonal language, will allow us to make definitions almost anywhere. However, there will be a time penalty when too many solutions are incorporated, as they will have to be repeatedly parsed by the interpreter during LSearch. To solve this problem, we add a hook called solution-corpus to the grammar for definition, which can be achieved in a similar way to previous-solution. However, then, the probability of defining *and* using a previous solution will greatly decrease. Assume that a previous solution is defined with a probability of $p_1$ and called with a probability of $p_2$. Since the grammar does not condition calling a previous solution on the basis of definition, the probability of a correct use is $p_1.p_2$; most of the time this logic will just generate semantically incorrect invocations of the past solutions. As an improvement, each definition is added to previous-solution as a production procedure which appends the function name to variables (e.g. for higher-order functions) and the solution number to the environment, and generates an empty production with a probability of zero when the solution has already been defined (the search algorithm must handle this case of course). This solution backtracks program search when multiple definitions of the same solution are encountered, avoiding generation of redundant programs. It is easy to make the probabilities of previous solutions modifiable. The newest solution is added with a probability of 0.5 and in later problems we allow the first update algorithm to modify the probabilities of solution productions.

### 5.3 Learning programming idioms

Programmers do not only learn of concrete solutions to problems, but they also learn abstract programs, or program schemas. One way to formalize this is that they learn sentential forms. If we can extract appropriate sentential forms, we can add these to the grammar, as well.

We construct the derivation tree from the leftmost derivation, with an obvious algorithm that we will omit. The current abstraction algorithm starts with the derivation sub-trees rooted at each expression in the current solution. For each derivation sub-tree, we prune the leaves from the bottom-up. At each pruning step, an abstract expression is output. The pruning algorithm works as follows: the tree `[Node <:S:> [Node <:B:> [Leaf bb]] [Node <:A:> [Leaf a]] [Node<:B:>[Leaf bbb]]]` is pruned one level to obtain `[Node <:S:> [Leaf <:B:>] [Leaf <:A:>] [Leaf <:B:>]]`. The pruning is iterated until a few symbols remain. Every abstract expression thus found is added to a new non-terminal that contains the abstract expressions of the current solution with equal probability. The new non-terminal is added to the top-level non-terminal abstract-expression with 0.5 probability, which is itself one of the productions for expression. These productions may later be modified and used by update algorithms one and two.

Note that the orthogonality of the language helps us in integrating programming idioms into HAM. Thus, several sentential forms are learnt from a single solution in this fashion corresponding to different syntactic abstractions. We think that the system will eventually learn complex programming idioms like recursion patterns and building data structures.

### 5.4 Frequent sub-program mining

Mining the solution corpus further enhances the guiding probability distribution. Frequent sub-programs in the solution corpus, i.e., sub-programs that occur with a frequency above a given support threshold, can be added again as alternative productions to the commonly occurring non-terminal expression in the Scheme grammar. For instance, if the solution corpus contains several `(lambda (x y) (* x y) )` subprograms the frequent sub-program mining would discover that and we can add it as an alternative expression to the Scheme grammar.

We would like to find all frequent subprograms that occur twice or more so that we can increase the probability of such sub-programs accordingly. We first interpret the problem of finding frequent sub-programs as a syntactic problem, disregarding semantic equivalences between sub-programs. Once formulated in our program representations of derivation trees as labelled rooted frequent subtree mining, the frequent sub-program mining algorithm is a reasonable extension of traditional frequent pattern mining algorithms. We have implemented a BFS patterned fast mining algorithm by exploiting the property that every sub-tree of a frequent tree is frequent. We find frequent sub-trees (with a support threshold of 2 currently) of all sub-trees of derivation trees rooted at expression in the solution corpus. At each update, a non-terminal hook frequent-expression in the grammar is rewritten by assigning probabilities according to the frequency of each frequent sub-program. Note that most frequent expressions are abstract.

## 6 Experiments

Our experimental tests were carried out at the TUBITAK ULAKBIM High Performance Computing Center on 144 AMD Opteron 6172 cores. We present detailed information about two training sequences to demonstrate the effectiveness of our update algorithms.

We know of no previous demonstration of realistic experiments over long training sequences. Solomonoff indicated that the goal of early training sequence design is to get problem solving information into the machine and complained that previous research has focused on parallel processing to solve difficult problems without adequately training their systems [13, Section 6]. In our future AGI system, we expect each HAM module to solve millions of small problems and learn from them, therefore the solution of long training sequences composed of easy/medium-level difficulty problems is much more realistic than running plain Levin Search on a hard problem with a few selected primitives. We can show the effectiveness of our memory system leaving no place for doubt through controlled experiments. We run the entire training sequence with updates turned off and on. If the update algorithms cause a speed-up over search with no update in a consistent fashion, we can infer that the update algorithms are effective. We use Conceptual Jump Size (CJS) to calculate the difficulty of a problem. $CJS = t_i/p_i$ where $t_i$ is the running time of solution program and $p_i$ is its a priori probability. The upper bound of Levin Search's running time is $2.CJS$ [13, Appendix A]. Our experiments may be preferable to calculating CJS's by hand, as in these experiments we are using Scheme R5RS in its full glory. Note that we are interested in only detecting whether any information transfer occurs across problems rather than trying to solve difficult problems with a machine that knows nothing but a universal computer.

**Table 1.** Performance of training sequence 0 with no update, $|HAM| = 17145$

| Problem | Time | # Trials | Errors | Cycles | Max Cyc. | $p_i$ | $t_i$ | CJS | $H(s_i)$ |
|---|---|---|---|---|---|---|---|---|---|
| inv. $f(x) = x$ | 2.60 | 1588 | 403 | 8404 | 318630 | 0.0277 | 15 | 540 | 5.16 |
| inv. $f(x) = 1/x$ | 4.83 | 9941 | 2455 | 131471 | $2.38 \times 10^6$ | $7.91 \times 10^-6$ | 20 | $2.52 \times 10^6$ | 16.94 |
| inv. $f(x) = \sqrt{x}$ | 9.57 | 8043 | 20968 | 1059997 | $2.38 \times 10^7$ | $8.79 \times 10^{-7}$ | 20 | $2.27 \times 10^7$ | 20.11 |
| all | 17.25 | | | | | | | | |

**Table 2.** Performance of training sequence 0 with update

| Time | # Trials | Errors | Cycles | Max Cyc. | $p_i$ | $t_i$ | CJS | $H(s_i)$ | $\lvert HAM \rvert$ |
|---|---|---|---|---|---|---|---|---|---|
| 2.53 | 1588 | 403 | 18404 | 318630 | 0.0277 | 15 | 540 | 5.16 | 17180 |
| 3.25 | 5038 | 1379 | 70501 | $1.19 \times 10^6$ | $1.38 \times 10^{-5}$ | 20 | $1.44 \times 10^6$ | 16.14 | 17231 |
| 4.28 | 12735 | 3930 | 176261 | $3.03 \times 10^6$ | $6.51 \times 10^{-6}$ | 20 | $3.07 \times 10^6$ | 17.23 | 17420 |
| 9.98 | | | | | | | | | |

We initially solved problems for inverting mathematical functions, for identity function, division, and square-root functions. Tables 1 and 2 show the performance of the system with no update and update respectively. For each problem in sequence 0, we give the time in seconds, number of trials, number of Scheme errors, number of Scheme execution cycles spent, number of maximum cycles allocated to search, a priori probability of solution ($p_i$), running time of solution in Scheme cycles ($t_i$), Conceptual Jump Size, the length of the implicit program code of the solution ($H(s_i) = -lg(p_i)$) and the size of HAM in bytes after solving the problem. Total time for the training sequence is also given. The initial time limit is $10^6$ cycles. It is seen that for the simplest problems, the HAM helps a bit, but not much: it results in less than 2 speedup for the entire training sequence. However, we also see that problem 3 has benefited the most, its $CJS$ has reduced almost 8-fold (corresponding to 3 bits of transfer from previous solutions), and it has doubled in speed.

We also developed a training sequence composed of operator induction problems. For each problem, we have a sequence of input and output pairs, and we approximate operator induction as described by Solomonoff [6,14]. Training sequence 1 contains, in order, the square function `sqr`, the addition of two variables `add`, a function to test if the argument is zero `is0`, all of which have 3 example pairs, fourth power of a number `pow4` with just 2 example pairs, boolean `nand`, and `xor` functions with 4 example pairs each. Tables 3 and 4 convey the performance of our system on training sequence 1 without update and with update, respectively.

**Table 3.** Performance of training sequence 1 with no update, $|HAM| = 17145$

| Problem | Time | # Trials | Errors | Cycles | Max Cyc. | $p_i$ | $t_i$ | CJS | $H(s_i)$ |
|---|---|---|---|---|---|---|---|---|---|
| sqr | 16.28 | $5.34 \times 10^5$ | $1.57 \times 10^5$ | $5.46 \times 10^6$ | $2.05 \times 10^8$ | $2.19 \times 10^{-7}$ | 37 | $1.68 \times 10^8$ | 22.12 |
| add | 19.9759 | $1.03 \times 10^6$ | $3.13 \times 10^5$ | $1.13 \times 10^7$ | $4.1 \times 10^8$ | $9.77 \times 10^{-8}$ | 40 | $4.09 \times 10^8$ | 23.28 |
| is0 | 7.57 | 41210 | 9531 | 430336 | $1.10 \times 10^7$ | $3.95 \times 10^{-6}$ | 34 | $8.59 \times 10^6$ | 17.94 |
| pow4 | 1759.45 | $3.34 \times 10^8$ | $1.38 \times 10^8$ | $3.24 \times 10^9$ | $2.55 \times 10^{11}$ | $1.67 \times 10^{-10}$ | 26 | $1.55 \times 10^{11}$ | 32.47 |
| nand | 3497.17 | $6.48 \times 10^8$ | $2.71 \times 10^8$ | $6.69 \times 10^9$ | $5.13 \times 10^{11}$ | $2.01 \times 10^{-10}$ | 56 | $2.78 \times 10^{11}$ | 32.21 |
| xor | 1848.8 | $3.38 \times 10^8$ | $1.3 \times 10^8$ | $3.54 \times 10^9$ | $2.53 \times 10^{11}$ | $2.01 \times 10^{-10}$ | 52 | $2.58 \times 10^{11}$ | 32.21 |
| all | 7150.06 | | | | | | | | |

**Table 4.** Performance of training sequence 1 with update

| Problem | Time | # Trials | Errors | Cycles | Max Cyc. | $p_i$ | $t_i$ | CJS | $H(s_i)$ | $\|HAM\|$ |
|---|---|---|---|---|---|---|---|---|---|---|
| sqr | 11.4 | $6.34 \times 10^5$ | $1.81 \times 10^5$ | $6.64 \times 10^6$ | $2.35 \times 10^8$ | $2.19 \times 10^{-7}$ | 37 | $1.68 \times 10^8$ | 22.12 | 17318 |
| add | 7.63 | $2.46 \times 10^5$ | $8.52 \times 10^4$ | $3.39 \times 10^6$ | $8.19 \times 10^7$ | $0.33 \times 10^{-6}$ | 40 | $1.19 \times 10^8$ | 21.5 | 17515 |
| is0 | 2.72 | 10202 | 2969 | 136363 | $2.14 \times 10^6$ | $0.13 \times 10^{-4}$ | 34 | $2.60 \times 10^6$ | 16.22 | 17566 |
| pow4 | 6.45 | $2.62 \times 10^5$ | $8.92 \times 10^4$ | $3.6 \times 10^6$ | $9.86 \times 10^7$ | $0.72 \times 10^{-6}$ | 54 | $7.39 \times 10^7$ | 20.38 | 17617 |
| nand | 209.53 | $2.55 \times 10^7$ | $1.12 \times 10^7$ | $3.72 \times 10^8$ | $1.51 \times 10^{10}$ | $0.50 \times 10^{-8}$ | 56 | $1.11 \times 10^{10}$ | 27.57 | 17962 |
| xor | 4.22 | 43749 | 14216 | 667625 | $1.18 \times 10^7$ | $0.47 \times 10^{-5}$ | 57 | $1.19 \times 10^7$ | 17.68 | 18438 |
| all | 245.1 | | | | | | | | | |

The overall speed-up of training sequence 1 with updates is 29.17 compared to the tests with no HAM update. This result indicates a consistent success of transfer learning in a long training sequence. The search time for the solutions in Table 4 tend to decrease compared to Table 3. The memory size has increased only 1293 bytes, for storing information for 6 operator induction problems, which corresponds to %7.5 increase in memory for 29.17 speed-up, which is a very favorable time-space trade-off. The solution of logical functions took longer than previous problems in Table 3, but we saw significant time savings in Table 4. Previous solutions are re-used aggressively. In Table 4, `pow4` solution `(define (pow4 x ) (define (sqr x ) (* x x)) (sqr (sqr x ) ))` re-uses the `sqr` solution and takes only $2.62 \times 10^6$ trials, its CJS speeds up 2097.4 times over the case with no update, and the search achieves 272 speed-up in running time.

## 7 Conclusion

We have described a stochastic CFG based incremental machine learning system in detail. We have introduced our realization of Solomonoff's Phase 1 machine and our Heuristic Algorithmic Memory (HAM) design. We have adapted R5RS Scheme as the reference universal computer to our system. The SCFG is used in parallel LSearch to calculate a priori probabilities and to generate programs efficiently avoiding syntactically incorrect programs. We derive sentences using leftmost derivation. We use parallel DFS algorithms for enumerating candidate programs. We have specialized productions for number literals, variable bindings and variable references.

We have proposed four update algorithms for incremental machine learning. All of them have been implemented and found to be fairly efficient. The effectiveness of our update logic has been demonstrated with experiments in one short and one long training sequence, a feat that has not been accomplished before to the best of our knowledge.

In the future, we plan to implement the Phase 2 of Solomonoff's Alpha system, and attempt integrating our AGI kernel to other AGI proposals such as the Gödel Machine [3], AIXI$^{tl}$ [15], as well as adapting new program search algorithms such as HSearch [2], and MOSES[16].

## References

1. Solomonoff, R.J.: A system for incremental learning based on algorithmic probability. In: Proceedings of the Sixth Israeli Conference on Artificial Intelligence, Tel Aviv, Israel (1989) 515–527
2. Hutter, M.: The fastest and shortest algorithm for all well-defined problems. International Journal of Foundations of Computer Science **13**(3) (2002) 431–443
3. Schmidhuber, J.: Ultimate cognition *à la* Gödel. Cognitive Computation **1**(2) (2009) 177–193
4. Schmidhuber, J.: Optimal ordered problem solver. Machine Learning **54** (2004) 211–256
5. R. Cilibrasi, P.V.: Clustering by compression. Technical report, CWI (2003)
6. Solomonoff, R.J.: Progress in incremental machine learning. In: NIPS Workshop on Universal Learning Algorithms and Optimal Search. (2002)
7. Solomonoff, R.J.: Algorithmic probability: Theory and applications. In Dehmer, M., Emmert-Streib, F., eds.: Information Theory and Statistical Learning, Springer Science+Business Media, N.Y. (2009) 1–23
8. Schmidhuber, J., Zhao, J., Wiering, M.: Shifting inductive bias with success-story algorithm, adaptive levin search, and incremental self-improvement. Machine Learning **28**(1) (1997) 105–130
9. Richard Kelsey, William Clinger, J.R.: Revised5 report on the algorithmic language scheme. Higher-Order and Symbolic Computation **11**(1) (1998) (editors).
10. Levin, L.A.: Universal sequential search problems. Problems of Information Transmission **9**(3) (1973) 265–266
11. John E. Hopcroft, Rajeev Motwani, J.U.: Introduction to Automata Theory, Languages, and Computation. Second edn. Addison Wesley (2001)
12. Merialdo, B.: Tagging english text with a probabilistic model. Computational Linguistics **20** (1993) 155–171
13. Solomonoff, R.J.: Algorithmic probability, heuristic programming and agi. In: Third Conference on Artificial General Intelligence. (2010) 251–157
14. Solomonoff, R.J.: Three kinds of probabilistic induction: Universal distributions and convergence theorems. The Computer Journal **51**(5) (2008) 566–570
15. Hutter, M.: Universal algorithmic intelligence: A mathematical top→down approach. In Goertzel, B., Pennachin, C., eds.: Artificial General Intelligence. Cognitive Technologies. Springer, Berlin (2007) 227–290

16. Looks, M.: Scalable estimation-of-distribution program evolution. In: Proceedings of the 9th annual conference on Genetic and evolutionary computation. (2007)

.